\documentclass{Interspeech}

\interspeechcameraready

\title{Sounding Like a Winner? Prosodic Differences in Post-Match Interviews}

\author[affiliation=]{Sofoklis}{Kakouros}
\author[affiliation=]{Haoyu}{Chen}

\affiliation[nocounter]{Center for Machine Vision and Signal Analysis}{University of Oulu}{Finland}
\email{sofoklis.kakouros@oulu.fi, haoyu.chen@oulu.fi}

\usepackage{comment}

\begin{document}

\maketitle

\begin{abstract}
    
    This study examines the prosodic characteristics associated with winning and losing in post-match tennis interviews. Additionally, this research explores the potential to classify match outcomes solely based on post-match interview recordings using prosodic features and self-supervised learning (SSL) representations. By analyzing prosodic elements such as pitch and intensity, alongside SSL models like Wav2Vec 2.0 and HuBERT, the aim is to determine whether an athlete has won or lost their match. Traditional acoustic features and deep speech representations are extracted from the data, and machine learning classifiers are employed to distinguish between winning and losing players. Results indicate that SSL representations effectively differentiate between winning and losing outcomes, capturing subtle speech patterns linked to emotional states. At the same time, prosodic cues—such as pitch variability—remain strong indicators of victory. 
\end{abstract}

\section{Introduction}

Although significant progress has been made in speech emotion recognition (SER) \cite{ahn2025multitask,kakouros2023speech,stafylakis2023extracting} and understanding \cite{rathi2024analyzing}, most research has focused on artificially designed tasks and subjective emotion annotations \cite{busso2008iemocap}. These tasks typically rely on predefined emotion categories (such as happiness, sadness, anger, etc.), and the emotion annotations are often influenced by the annotators' personal emotional understanding and cultural background \cite{livingstone2018ryerson}. While these approaches are effective, they are limited in their application to more complex and natural environments.

In this work, we propose a novel speech emotion recognition task in a natural setting: predicting the outcome of a tennis match based on the post-match interview speech of the athletes. We then leverage this task for prosodic analysis and inference. This task has several innovative aspects: 1. Emotion recognition in real-world scenarios: Unlike artificially constructed emotion datasets, our study uses speech data from real post-match interviews, which capture the athletes' genuine emotional expressions in high-pressure situations, offering more natural and diverse emotional cues. 2. Emotion inference without explicit emotion labels: In contrast to traditional emotion recognition tasks, our study does not rely on predefined emotion labels. Instead, we investigate whether the athlete won or lost the match based on acoustic features such as pitch, speech rate, and intensity. This task challenges the model to detect subtle emotional shifts rather than simply categorizing emotions. 3. Direct connection between emotion and behavior (match outcome): Our task links emotional expression to a specific behavior (match result), aiming to explore the relationship between emotional states and actual actions. This not only provides a fresh perspective on emotion recognition but also offers new methods and insights for fields such as sports psychology and athlete behavior analysis.

In summary, this work introduces an unprecedented natural-scenario task for speech emotion recognition and understanding, leveraging a dataset that captures speech in winning and losing scenarios. By adopting this more naturalistic approach we explore the acoustic-prosodic differences between these conditions, analyzing how features such as pitch and intensity vary depending on the speaker's outcome in the competitive and emotionally charged situations of tennis matches. To assess the accuracy of emotion inference in this task, we employ state-of-the-art self-supervised learning techniques, opening new avenues for research in emotion computing.

\subsection{Emotion recognition}

The challenges in SER are diverse and can generally be divided into three main areas. The first challenge involves developing representations that effectively capture the acoustic variations that signal different emotions. Traditionally, this has relied on handcrafted features such as mel-frequency cepstral coefficients (MFCCs), filter banks (FBs) \cite{ververidis2006emotional}, and prosodic feature sets like the extended Geneva Minimalistic Acoustic Parameter Set (eGeMAPS) \cite{pepino2021emotion}. However, the rise of self-supervised learning (SSL) techniques, including HuBERT \cite{hsu2021hubert}, Wav2vec 2.0 \cite{baevski2020wav2vec}, and WavLM \cite{chen2021wavlm}, has introduced more refined representations that have achieved state-of-the-art results in SER \cite{kakouros2023speech, yang2021superb, pepino2021emotion}. These methods represent a shift toward machine-driven learning of speech representations directly from raw audio, enabling a more sophisticated and layered analysis of emotional expression in speech.

The second challenge lies in modeling the temporal dynamics of emotions, which can manifest over both short and extended speech sequences. While emotions can be conveyed within single words or short phrases, they can also span multiple utterances, requiring analytical approaches capable of capturing and linking emotional cues across varying time scales. Solutions range from simple statistical pooling of time-based self-supervised representations \cite{yang2021superb} to more advanced sequence modeling techniques, such as recurrent neural networks (RNNs), which aim to map and interpret the emotional content of speech \cite{sarma2018emotion}. The complexity of SER is further amplified by the inherent ambiguity of emotional expressions, which are often influenced by multimodal factors such as facial expressions and body language \cite{kim2015leveraging, mower2009interpreting}. This ambiguity frequently leads to misinterpretation, even among human evaluators, underscoring the importance of SER models that can disambiguate overlapping and uncertain emotional cues. To tackle this challenge, researchers have proposed various methods, including specialized loss functions \cite{liu2021speech} and label smoothing techniques to refine target labels \cite{kakouros2023speech}.

\subsection{Prosodic analysis of emotions}

\begin{figure*}[t]
  \centering
  {\includegraphics[width=0.68\textwidth]{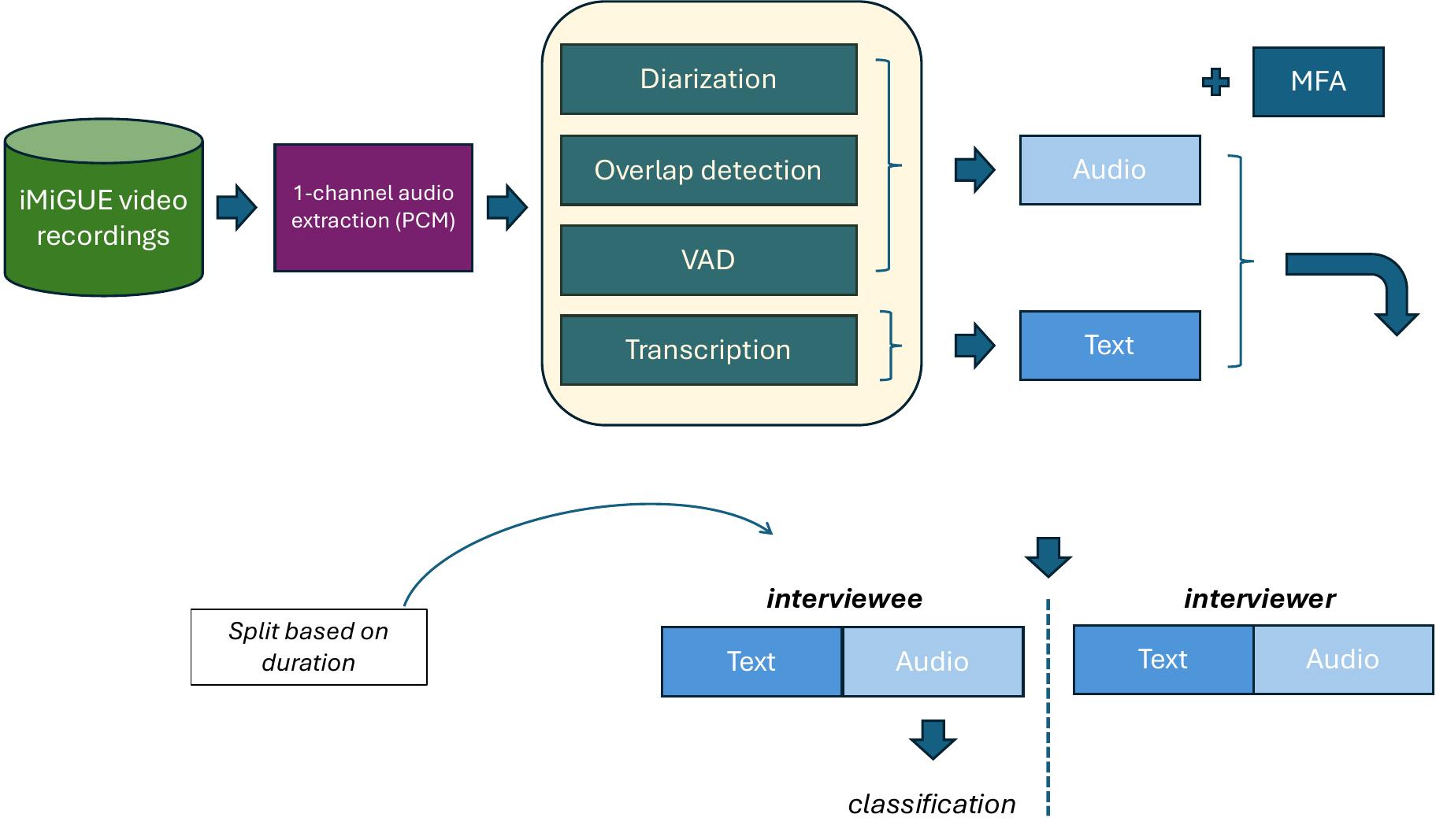}}
  \caption{Overview of the processing pipeline for audio extraction, feature extraction, and classification}
  \label{fig:pipeline_overview}
\end{figure*}

A fundamental aspect of emotions is encoded in their acoustic realization. Consequently, acoustic-prosodic analysis plays a crucial role in understanding how variations in speech signals correspond to different emotional states. Prosodic features—such as pitch, intensity, speaking rate, and rhythm—are widely recognized as strong indicators of emotional expression \cite{scharenborg2018effect}. For instance, heightened pitch and intensity are often associated with emotions like anger or excitement, whereas lower pitch and reduced intensity typically signal sadness or calmness \cite{eyben2015geneva, schuller2011recognising}. Similarly, variations in speech rate and articulation patterns help distinguish emotions; for example, speech tends to slow down and become more drawn out when expressing sadness, while it is generally faster for emotions such as happiness or fear \cite{gobl2003role, xu2013speech}. These prosodic variations serve as valuable cues for automatic emotion recognition systems, enabling them to differentiate between subtle emotional states beyond lexical content.

Beyond traditional prosodic measures, recent studies have investigated higher-order acoustic-prosodic patterns that capture the dynamic nature of emotional speech. Temporal variations in pitch and intensity contours, spectral balance, and voice quality (e.g., breathiness or roughness) have been shown to influence emotion perception \cite{gobl2003role, lee2002combining}. Additionally, research highlights that prosodic patterns can be culture- and language-dependent, affecting how emotions are conveyed and interpreted \cite{latif2020speech}. In the context of this study, analyzing acoustic-prosodic differences between winning and losing speech may uncover distinctive patterns that characterize success-related vocal expressions. Identifying these patterns can offer deeper insights into the emotional and psychological impact of winning versus losing, contributing to a more nuanced understanding of emotional variation in speech.

Rather than focusing on improving speech emotion recognition (SER) performance or developing a new SER system, this study examines the potential of SSL representations in capturing and distinguishing between positive and negative emotions in a novel dataset. Additionally, this work explores the acoustic-prosodic characteristics of winning versus losing speech, aiming to answer the question: What prosodic features distinguish winning speech from losing speech?

\section{Materials and data processing}

\subsection{iMiGUE dataset}

The iMiGUE (Identity-free Micro-Gesture Understanding and Emotion Analysis) dataset is a specialized video dataset designed for studying micro-gestures and their relationship to emotional states while ensuring privacy \cite{liu2021imigue}. Unlike conventional emotion recognition datasets that focus on facial expressions or speech, iMiGUE captures subtle, unintentional body movements that reflect internal emotions. The dataset consists of 359 videos from Grand Slam tennis post-match press conferences, featuring 72 players (36 male, 36 female) from 28 countries. Each video is annotated with micro-gesture categories at the clip level and emotional states (positive or negative) at the video level, based on match outcomes. Importantly, it is identity-free: while faces are masked or removed to protect personal identities, voices are preserved.

While iMiGUE provides a rich set of 18,499 annotated micro-gesture samples across 32 distinct categories, this study explores a novel perspective by focusing on the audio modality. Specifically, the vocal expressions from the 359 post-match interviews conducted in English are analyzed, investigating how speech patterns correlate with emotional states. This approach offers new insights into the relationship between vocal and nonverbal emotional cues.

A human evaluation of the recordings, involving subjective annotations, was initially planned in this study; however, the decision was ultimately made to omit it. This was because it would create an unfair comparison. Human evaluators inherently understand the semantic content of the interviews, making it significantly easier for them to interpret match outcomes (positive/negative). For example, inclusion of the textual content (based on transcripts) using LLaMA-3.1-8B was tested, and achieved extremely high accuracy—over 90\%—demonstrating the substantial advantage humans (and content-aware models) have in this task.

\subsection{Data pre-processing}
The challenge in the iMiGUE dataset was to extract the audio segments where the athlete was actively responding to questions or speaking in general. Since the iMiGUE dataset consists of manually labeled gesture annotations for the athletes, it does not include speaker turn annotations or information on who is speaking. To address this, the pipeline illustrated in Figure \ref{fig:pipeline_overview} provides an overview of the entire process.

First, audio is extracted from each video recording. Then, speaker diarization, overlap detection, and voice activity detection (VAD) are applied using Pyannote. Transcriptions are generated using Whisper Large, with the language flag explicitly set to extract transcripts in English. All extracted annotations—including speaker labels, overlap segments, VAD markers, and transcriptions—are stored in TextGrid files, each in separate layers.

Additionally, force alignment is performed using Montreal Forced Aligner (MFA) to determine precise word boundaries, which are saved in a separate TextGrid file. To distinguish the athlete from the journalists, total speaker durations are computed, and the speaker with the longest speech duration is identified as the athlete. The corresponding speech segments are then extracted from the full audio recording and saved separately. Each extracted segment for the athlete is indexed sequentially based on its position in the original recording. While the original recordings can last several minutes, the extracted segments are typically only a few seconds long. These segments are then organized separately for the athlete (interviewee) and the journalists (interviewers), ensuring clear differentiation between the two.

\section{Method}

\subsection{Feature extraction}
A standard set of acoustic features was extracted using eGeMAPS \cite{eyben2015geneva}, which includes 88 distinct features and functionals for each interviewee segment. These features encompass the means and standard deviations of fundamental frequency (F0), loudness, spectral tilt, and mel-frequency cepstral coefficients (MFCCs). For each selected speech segment, all 88 eGeMAPS features and functionals were computed, resulting in a single feature vector representing the entire segment.

To further assess the separability of emotion classes, additional features were extracted using state-of-the-art self-supervised learning (SSL) representations. SSL has driven advancements across various speech technology domains, including prosody modeling. Compared to standard acoustic features like MFCCs, SSL representations have significantly higher dimensionality, providing a richer representation space and greater potential for capturing relevant speech characteristics.

SSL features were extracted using \texttt{Wav2vec 2.0 large} \cite{baevski2020wav2vec}, \texttt{HuBERT large} \cite{hsu2021hubert}, and \texttt{exHuBERT} \cite{amiriparian2024exhubert}. ExHuBERT enhances HuBERT by extending its architecture and fine-tuning it on 37 emotion datasets to improve multilingual speech emotion recognition. All model checkpoints were sourced from the \texttt{Hugging Face} model library. Representations were obtained from the final transformer layer of the pre-trained models. To generate an utterance-level representation, feature-level embeddings were pooled by computing the mean over each speech segment, resulting in a 1024-dimensional vector per utterance. The pooled representations were then averaged across all individual segments corresponding to each recording. As a result, the athlete's speech from each recording in the dataset was represented by a single embedding.

\subsection{Supervised learning}
For supervised classification, a neural network was used to distinguish between "win" and "lose" categories. The neural network consists of three hidden layers, each incorporating batch normalization for stable training, LeakyReLU activation for non-linearity, and dropout regularization (0.3) in the first two layers to prevent overfitting. The first hidden layer maps the input features to 256 dimensions, followed by a reduction to 128 and then 64 dimensions in the subsequent layers. The final output layer projects the feature space to two dimensions, corresponding to the classification labels. CrossEntropyLoss is used as the loss function.

Optimization is handled using Adam optimizer with a learning rate of $1 \times 10^{-3}$ while a StepLR scheduler reduces the learning rate by a factor of 0.5 every five epochs to improve convergence. Training is conducted with a batch size of 16. To address class imbalance (roughly 80\% of the data belong to the win class), data augmentations are used. In particular, the Synthetic Minority Over-sampling Technique (SMOTE) is utilized \cite{chawla2002smote}. SMOTE is a data augmentation technique used to balance imbalanced datasets by generating synthetic samples for the minority class. Instead of simply duplicating existing samples, SMOTE creates new instances by interpolating between existing ones.

\subsection{Shapley feature explanations}
To determine feature importance for OpenSMILE features, Shapley explanations are applied. Shapley feature explanations, as implemented in the python SHAP (SHapley Additive exPlanations) library, provide a robust way to interpret the contributions of each feature in a machine learning model's predictions. SHAP values quantify the marginal contribution of each input feature by considering all possible feature coalitions, ensuring a globally consistent and locally accurate explanation of the model's behavior \cite{lundberg2017unified,vstrumbelj2014explaining}. The goal with SHAP is to identify which prosodic features have the greatest influence on the classifier's decision when distinguishing between the win and lose classes.

\section{Experiments}

The dataset was divided into training (70\%), validation (20\%), and test (10\%) sets, ensuring that each subset contained distinct speakers with no overlap to prevent data leakage and improve generalization.

\subsection{Prosodic analysis}
For the prosodic analysis, a neural network was trained for 50 epochs using 88-dimensional openSMILE features, which capture a variety of acoustic and prosodic characteristics. After training, the test set was used to compute SHAP values, providing insights into feature importance and model interpretability.

\subsection{Win/Lose classification}
For the classification experiments, the classifier was trained separately for 50 epochs on the same data split using different feature representations: Wav2Vec 2.0, HuBERT, exHuBERT, and openSMILE features. The performance of each approach was evaluated to compare their effectiveness in the classification task.

\section{Results and analysis}
\begin{table}[t]
    \centering
    \caption{Classification performance (\%) for different SSL representations and openSMILE.}
    \label{tab:performance}
    \begin{tabular}{lcccc}
        \toprule
        & \textit{ACC} & \textit{PRC} & \textit{RCL} & \textit{F1} \\
        \midrule
        HuBERT & 65.9 & 65.3 & 61.1 & 55.7 \\
        exHuBERT  & 60.7 & 60.6 & 59.7 & 56.3 \\
        Wav2Vec 2.0 & 63.9 & 63.7 & 62.5 & 60.0 \\
        openSMILE  & 44.4  & 44.9  &  44.4 & 37.2 \\
        \bottomrule
    \end{tabular}
\end{table}

\subsection{Prosodic analysis}
\begin{figure}[b]
  \vspace{-4mm}
  \centering
  \includegraphics[width=\linewidth]{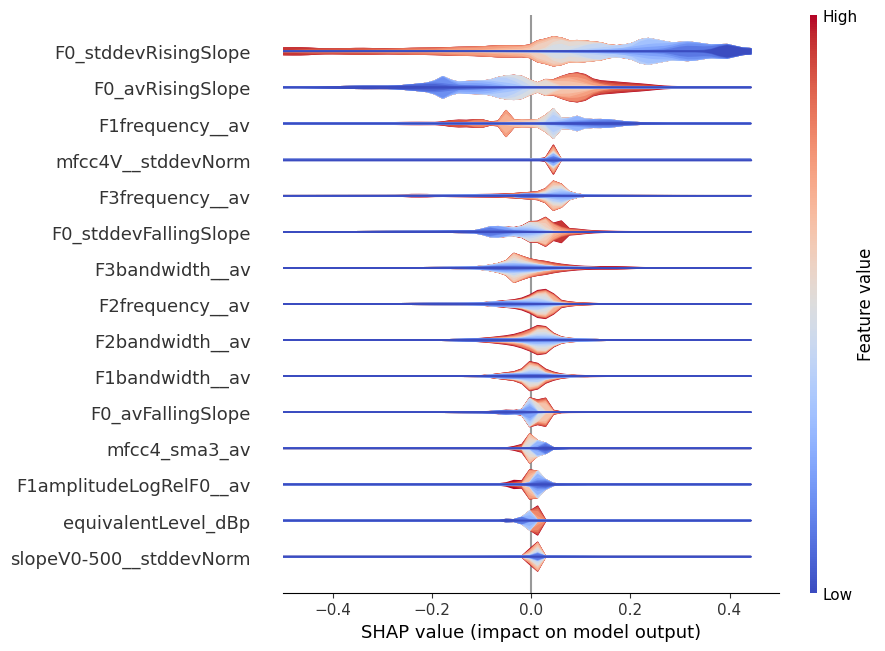}
  \caption{SHAP OpenSMILE feature importance values.}
  \label{fig:shap_summary}
  \vspace{-4mm}
\end{figure}

Our findings show (see Figure \ref{fig:shap_summary}) that winners tend to have more dynamic pitch movements, varied spectral features, and higher intensity, whereas players who lost their match exhibit more monotonous speech patterns with less variation in frequency and intensity. This pattern is particularly evident in features related to pitch slope variability and rising pitch trends, such as $F0\_stddevRisingSlope$ and $F0\_avRisingSlope$.

Specifically, the $F0\_stddevRisingSlope$, which captures the variability in the rising segments of pitch contours, seems to play an important role in distinguishing winners from athletes who lost the match. A higher standard deviation indicates greater fluctuations in pitch slopes, reflecting more expressive and dynamic speech. Winners exhibit high and low values of this feature, suggesting that their speech contains a wide range of variations in rising pitch slopes. In contrast, athletes who lost the match tend to have uniformly high values, indicating more unstable pitch slopes. This instability may contribute to less stable and more variable speech patterns.

Correspondingly, $F0\_avRisingSlope$, which measures the average steepness of rising pitch slopes, highlights another key difference in speech dynamics. The differences in $F0\_stddevRisingSlope$ and $F0\_avRisingSlope$ together suggest that winning speakers use more dynamic and expressive pitch changes, whereas losing speakers exhibit more monotonous and unvaried intonation patterns. Another observation is with respect to the $equivalentLevel\_dBp$ (overall sound level in decibels). The SHAP values indicate that higher sound level values are associated with win.

\subsection{Win/Lose classification}
Table \ref{tab:performance} presents the classification performance of different SSL representations and the openSMILE feature set across four evaluation metrics: accuracy (ACC), precision (PRC), recall (RCL), and F1-score (F1).

Among the SSL-based models, HuBERT achieves the highest accuracy (65.9\%) and precision (65.3\%), though its recall (61.1\%) and F1-score (55.7\%) are slightly lower. Wav2Vec 2.0 follows closely with 63.9\% accuracy, 63.7\% precision, and the highest recall (62.5\%), leading to the highest F1-score (60.0\%). exHuBERT performs slightly worse, with 60.7\% accuracy and lower recall (59.7\%), but has a relatively balanced precision (60.6\%) and F1-score (56.3\%).

In contrast, openSMILE significantly underperforms compared to the SSL models, achieving the lowest accuracy (44.4\%), recall (44.4\%), and F1-score (37.2\%), with only a marginally higher precision (44.9\%). This suggests that handcrafted acoustic features extracted from openSMILE are less effective for the classification task compared to SSL-based representations.

Overall, HuBERT and Wav2Vec 2.0 demonstrate the best performance, with HuBERT achieving the highest accuracy and precision, while Wav2Vec 2.0 excels in recall and F1-score. The results highlight the advantage of SSL models over traditional feature extraction approaches like openSMILE.

Interestingly, the exHuBERT model, despite being fine-tuned on a diverse set of emotion corpora, performs worse than the pre-trained HuBERT and Wav2Vec 2.0 models. Although exHuBERT is derived from HuBERT Large, the additional emotional fine-tuning does not enhance performance; instead, it appears to degrade it. This degradation could be attributed to a trade-off between emotional sensitivity and task-specific discriminability. Fine-tuning on emotion corpora may have adapted exHuBERT to capture nuanced affective expressions which are potentially different from the acoustic cues most relevant for win/lose classification. Future work could explore domain-specific fine-tuning strategies or multi-task learning approaches to balance emotional awareness with the primary classification task.

\section{Conclusions}

This paper presents a prosodic analysis and win/lose classification performance on a novel dataset of post-match interview recordings. The acoustic-prosodic analysis reveals that winners exhibit more dynamic pitch movements, varied spectral features, and greater intensity. In contrast, players who lost their match tend to have flatter, more monotonous speech patterns with smaller F0 and intensity variation. Regarding classification performance, while traditional acoustic-prosodic features provide valuable insights into how athletes speak, they underperform compared to self-supervised learning (SSL) representations extracted from Wav2Vec 2.0 and HuBERT. These findings highlight the advantages of SSL-based approaches over handcrafted acoustic features for distinguishing between winning and losing players based on their speech patterns.

\section{Acknowledgements}

\ifinterspeechfinal
     This work was supported by the University of Oulu and the Research Council of Finland, PROFI7 352788 project. The authors wish to acknowledge CSC – IT Center for Science, Finland, for providing the computational resources.
\else
     Acknowledgments will be added
\fi

\bibliographystyle{IEEEtran}
\bibliography{mybib}

\end{document}